\documentclass[letterpaper, 10 pt, conference]{ieeeconf}  

\IEEEoverridecommandlockouts                              

\usepackage{amsmath} 
\usepackage{amssymb}  
\usepackage{booktabs} 
\usepackage{graphicx}
\usepackage{subcaption}
\title{\LARGE \bf
Image Frame Dynamic Object Segmentation and Ego Motion Estimation using Radar Image Fusion
}

 \author{Astik Srivastava, Suhani Grover, Avinash Sharma and Madhava Krishna}

\begin{document}

\maketitle
\thispagestyle{empty}
\pagestyle{empty}

\begin{abstract}

Dynamic object segmentation and ego-motion estimation are closely coupled
problems in autonomous driving, as accurate ego-motion estimation typically
requires static scene observations, while identifying static observations
requires knowledge of the ego motion. We present Radar-Dot, a radar--RGB
framework that exploits radar Doppler measurements to address this coupling.
Radar returns are first used to estimate ego velocity through a linear Doppler
constraint, with residual-based static/dynamic segmentation and robust
estimation used to reduce the influence of moving objects. The estimated
motion is then combined with metric depth and dense optical flow to identify
image regions whose observed motion is inconsistent with the rigid scene
motion. Experiments on 10 nuScenes scenes (part of nuscenes-mini) demonstrate that the resulting geometric pipeline achieves 20.24\% dynamic IoU and 33.67\%
F1-score over 394 frame pairs, while radar-based static-point filtering
improves ego-velocity estimation compared with using all radar returns.
These results demonstrate the potential of radar as a modality for jointly improving ego-motion estimation and dynamic
object segmentation.

\end{abstract}

\section{Introduction}\label{Sec:Introduction}
Segmenting and tracking dynamic objects is extremely important for autonomous driving. This problem becomes complicated because of its circular dependence with the task of ego motion estimation. Existing methods for segmenting out dynamic objects in 3d \cite{xu2023onboard}\cite{xu2025lv} rely on accurate pose and velocity estimates of the ego. Traditional methods for finding ego motion using onboard sensors rely on Lidar or Visual Odometry or a combination of the two, depending on sensors available on the vehicle. Lidar Odometry uses ICP whereas Visual Odometry relies on epipolar geometry to find transformation between consequtive frames, which can be used to estimate ego motion. Both these methods rely on the assumption that the environment other than ego is stationary. This constructs a circular problem, where accurate motion estimation is required for segmenting dynamic obstacles, but knowing which features are static (and hence can be trusted for extracting ego motion from geometry) and which features are dynamic becomes important for accurate ego motion estimation. Adding Radars to a modality offers a unique advantage to solve this circular problem. Typical automative radars are capable of providing doppler velocity of the point of return along with its position in 2 dimensions (3 dimensions if equipped with a 4D radar). The doppler velocity of multiple radar returns along with the information about a starting initial velocity of the ego (which is generally 0 m/s) can be used to estimate ego motion. There are multiple works which show the application of this concept \cite{abu2023radar}. Doppler velocity can also be used to find the velocity of other objects once the ego velocity is known. \cite{zhu2026redefining} presents a strategy for simultaneously segmenting radar pointcloud into static and dynamic and also estimate Ego velocity. They however do not translate the segmentation information to any rgb image.
In this paper, we present following contributions:
\begin{itemize}
    \item We estimate ego velocity from radar Doppler measurements while explicitly separating static and dynamic radar returns.
    \item We use the radar-derived ego velocity to predict rigid background flow and combine flow and depth consistency to identify independently moving image regions.
    \item We evaluate the complete closed-loop pipeline on ten nuScenes scenes using nuInsSeg-based dynamic-object masks and analyze the contributions of radar ego-motion and the flow/depth consistency formulation.
\end{itemize}

\section{Methodology}
\subsection{Radar Odometry Estimation}\label{sec:radar_odometry}
As mentioned in \ref{Sec:Introduction}, Radar measurements can be used to estimate ego velocity. Assuming $N$ radar point returns, each containing values $(r, \theta, d)$, where $r$ indicates the range, $\theta$ indicates the azimuthal position of the return and $d$ indicates the line of sight velocity, a simple linear system of equation can be constructed to find the ego velocity $(v_x, v_y)$ that best explains the LoS velocities of each of the radar returns under a static environment assumption, since each radar return needs to satisfy:
\begin{equation}
    d_i = \frac{-\vec{v}_{ego}\cdot\vec{P_i}}{\lVert \vec{P_i} \rVert}
\end{equation}
Here, $\vec{P} = (r\cos(\theta), r\sin(\theta))$ represents the position of radar return in cartesian coordinates for the $i^{th}$ radar return. Defining the corresponding unit bearing vector as:

\begin{equation}
\vec{u}_i =
\frac{\vec{P}_i}{\lVert\vec{P}_i\rVert}
=
\begin{bmatrix}
\cos\theta_i\
\sin\theta_i
\end{bmatrix},
\end{equation}

The Doppler constraint for each radar return can be written as:

\begin{equation}
\begin{bmatrix}
\cos\theta_i & \sin\theta_i
\end{bmatrix}
\begin{bmatrix}
v_x\
v_y
\end{bmatrix}
= -d_i.
\end{equation}

Stacking the constraints from all $N$ radar returns gives the following linear system:

\begin{equation}
    \underbrace{
    \begin{bmatrix}
        \cos\theta_i & \sin\theta_i
    \end{bmatrix}
    }_{\mathbf{H}_i}
    \underbrace{
    \begin{bmatrix}
        v_x\\
        v_y
    \end{bmatrix}
    }_{\vec{v}_{ego}}
    = -d_i,
    \qquad i=1,\ldots,N.
\end{equation}
Since the system is generally overdetermined, the ego velocity is estimated using a least-squares formulation:
\begin{equation}
    \hat{\mathbf{v}}_{ego}
    =
    \underset{\mathbf{v}}{\operatorname{argmin}}
    \left\|
        \mathbf{H}\mathbf{v}-\mathbf{b}
    \right\|_2^2.
\end{equation}
The estimated ego velocity is integrated over time to reconstruct the vehicle trajectory. Since the radar formulation estimates only translational velocity and does not estimate heading, the ground-truth rotation $\mathbf{R}_t$ is used to transform the ego-frame velocity into the world frame:

\begin{equation}
    \hat{\mathbf{v}}_{\mathrm{world},t}
    =
    \mathbf{R}_t
    \hat{\mathbf{v}}_{\mathrm{ego},t}.
\end{equation}
Starting from the ground-truth initial position $\mathbf{p}_0$, the trajectory is recursively integrated as
\begin{equation}
    \hat{\mathbf{p}}_{t+1}
    =
    \hat{\mathbf{p}}_t
    +
    \mathbf{R}_t
    \hat{\mathbf{v}}_{\mathrm{ego},t}
    \Delta t_t.
\end{equation}

The static-scene assumption, however, is violated by radar returns originating
from moving objects. Such returns exhibit an additional velocity component due
to the motion of the reflecting object and therefore do not satisfy the ego
motion constraint exactly. To identify these returns, the estimated ego
velocity can be used to predict the Doppler velocity expected for each radar
point. Given an estimated ego velocity $\hat{\mathbf{v}}_{ego}$, the predicted
line-of-sight velocity of the $i^{th}$ radar return is

\begin{equation}
    \hat{d}_i =
    -\mathbf{u}_i^\top \hat{\mathbf{v}}_{ego}.
\end{equation}

The Doppler residual is then defined as the difference between the measured
and predicted line-of-sight velocities:

\begin{equation}
    e_i =
    \left|d_i-\hat{d}_i\right|
    =
    \left|d_i+\mathbf{u}_i^\top\hat{\mathbf{v}}_{ego}\right|.
\end{equation}

For a static radar return, the measured Doppler velocity is expected to be
well explained by the ego motion model, resulting in a small residual. In
contrast, a return originating from a moving object generally produces a
larger residual because its measured Doppler velocity contains both the
contribution from ego motion and the object's own motion. The radar returns
can therefore be segmented into static and dynamic points using a residual
threshold $\tau$:

\begin{equation}
    \mathcal{S}
    =
    \left\{
    i \mid e_i < \tau
    \right\},
    \qquad
    \mathcal{D}
    =
    \left\{
    i \mid e_i \geq \tau
    \right\},
\end{equation}

where $\mathcal{S}$ and $\mathcal{D}$ denote the sets of static and dynamic
radar returns, respectively.

The segmentation can be incorporated into the ego-velocity estimation through
an iterative procedure. First, an initial velocity estimate is obtained using
all available radar returns. The Doppler residual of every return is then
computed using this estimate, and returns whose residual exceeds the
threshold $\tau$ are classified as dynamic. A second velocity estimate is
subsequently obtained using only the points classified as static:

\begin{equation}
    \hat{\mathbf{v}}_{ego}^{\,static}
    =
    \underset{\mathbf{v}}{\operatorname{argmin}}
    \sum_{i\in\mathcal{S}}
    \left(
        \mathbf{u}_i^\top\mathbf{v}+d_i
    \right)^2.
\end{equation}

This procedure reduces the influence of independently moving objects on the
ego-velocity estimate. The resulting static-point estimate can be used both
for radar odometry and as a basis for identifying dynamic radar returns. In
practice, a robust estimator such as RANSAC can additionally be applied to
the Doppler constraints to reduce the sensitivity of the initial velocity
estimate to dynamic objects and other radar outliers.

The resulting segmentation provides a direct correspondence between radar
Doppler measurements and the estimated ego motion. Points whose measured
Doppler is consistent with the estimated ego velocity are treated as
observations of the static environment, whereas points exhibiting
substantial Doppler inconsistency are treated as potential dynamic
observations. This segmentation is subsequently used by the dynamic-object
detection pipeline described in \ref{sec:dynamic_segmentation}.

\subsection{Dynamic Object Segmentation}\label{sec:dynamic_segmentation}
\subsubsection{RGB Mask Generation}
Given the estimated ego velocity at time $t-1$, the RGB images and radar measurements from two consecutive frames, $I_{t-1}, I_t$ and $R_{t-1}, R_t$, are used to identify image regions whose motion is inconsistent with the estimated camera motion. The key assumption is that pixels belonging to static scene elements should exhibit optical flow that is consistent with the rigid motion of the camera, whereas independently moving objects will produce a residual motion.

First, metric depth maps are estimated for both frames using the radar-RGB depth network, denoted by $D_{t-1}$ and $D_t$. For this we use JustDepth\cite{yun2026justdepth} to get a metric depth map using RGB and Radar pointcloud.
The estimated ego velocity is then used to construct the relative camera motion between the two frames. Since the radar formulation provides translational velocity but does not estimate heading, the ground-truth heading is used only to obtain the rotational component of the camera motion. The translational component is obtained from the radar velocity as

\begin{equation}
    \mathbf{t}_{t-1\rightarrow t}
    =
    \mathbf{v}_{\mathrm{ego},t-1}
    \Delta t.
\end{equation}

Together with the corresponding rotation $\mathbf{R}_{t-1\rightarrow t}$, this defines the relative camera transformation

\begin{equation}
    \mathbf{T}_{t-1\rightarrow t}
    =
    \begin{bmatrix}
        \mathbf{R}_{t-1\rightarrow t} &
        \mathbf{t}_{t-1\rightarrow t}\\
        \mathbf{0}^{T} & 1
    \end{bmatrix}.
\end{equation}

For each pixel $\mathbf{p}=(u,v)^T$ in $I_{t-1}$, its depth $D_{t-1}(u,v)$ is used to back-project it into 3D using the camera intrinsic matrix $\mathbf{K}$:

\begin{equation}
    \mathbf{X}_{t-1}
    =
    D_{t-1}(u,v)
    \mathbf{K}^{-1}
    \begin{bmatrix}
        u\\
        v\\
        1
    \end{bmatrix}.
\end{equation}

Under the assumption that the corresponding scene point is static, its position in frame $t$ can then be predicted as

\begin{equation}
    \mathbf{X}_{t}
    =
    \mathbf{R}_{t-1\rightarrow t}\mathbf{X}_{t-1}
    +
    \mathbf{t}_{t-1\rightarrow t}.
\end{equation}

Projecting $\mathbf{X}_{t}$ back onto the image plane gives the expected pixel location $(u',v')$:

\begin{equation}
    \begin{bmatrix}
        u'\\
        v'\\
        1
    \end{bmatrix}
    \sim
    \mathbf{K}\mathbf{X}_{t}.
\end{equation}

The expected rigid optical flow is therefore

\begin{equation}
    \mathbf{f}_{\mathrm{est}}(u,v)
    =
    \begin{bmatrix}
        u'-u\\
        v'-v
    \end{bmatrix}.
\end{equation}

In parallel, dense optical flow between $I_{t-1}$ and $I_t$ is estimated using a pretrained optical-flow network, providing the observed flow $\mathbf{f}_{\mathrm{obs}}$. For static scene points, $\mathbf{f}_{\mathrm{obs}}$ should agree with $\mathbf{f}_{\mathrm{est}}$. Thus, their disagreement provides a cue for independently moving objects. The flow residual is defined as

\begin{equation}
    e_{\mathrm{flow}}(u,v)
    =
    \left\|
        \mathbf{f}_{\mathrm{obs}}(u,v)
        -
        \mathbf{f}_{\mathrm{est}}(u,v)
    \right\|_2.
\end{equation}

However, flow residual alone can be ambiguous, particularly for objects moving along the viewing direction where the induced image motion can be small. Since metric depth is available in both frames, an additional depth-consistency constraint is used. The depth predicted by rigidly transforming the point from frame $t-1$ is compared with the depth estimated independently at its projected location in frame $t$:

\begin{equation}
    e_{\mathrm{depth}}(u,v)
    =
    \left|
        D_t(u',v') - Z_{\mathrm{est}}
    \right|,
\end{equation}

where $Z_{\mathrm{est}}$ is the third component of $\mathbf{X}_t$. Static points are expected to have small depth residuals, while independently moving points can violate this constraint.

The two residuals are combined into a normalized dynamic score:

\begin{equation}
    S(u,v)
    =
    \frac{e_{\mathrm{flow}}(u,v)}
    {\sigma_{\mathrm{flow}}(u,v)}
    +
    \lambda
    \frac{e_{\mathrm{depth}}(u,v)}
    {\sigma_{\mathrm{depth}}(u,v)},
\end{equation}

where $\sigma_{\mathrm{flow}}$ and $\sigma_{\mathrm{depth}}$ are tolerance factors and $\lambda$ controls the contribution of the depth consistency term. These terms are defined as:
\begin{equation}
\begin{aligned}
\sigma_{\mathrm{flow}}(u,v)
&=
\left(1+
\frac{\widetilde{Z}}
{\max(D_{t-1}(u,v),Z_{\min})}\right) \\
&\quad\times
\left(1+
\exp\left[
-\frac{d_{\mathrm{FOE}}(u,v)^2}
{2\sigma_{\mathrm{FOE}}^2}
\right]\right).
\end{aligned}
\label{eq:sigma_flow}
\end{equation}
where $\widetilde{Z}$ is the median scene depth and
$d_{\mathrm{FOE}}$ is the distance to the focus of expansion. The
range and FOE terms increase tolerance where flow is less reliable.
The depth tolerance is defined as
\begin{equation}
\sigma_{\mathrm{depth}}(u,v)
=
\max(D_{t-1}(u,v),Z_{\min}),
\end{equation}
providing range-dependent normalization of the depth residual.
Finally, a robust frame-wise threshold is obtained using the median and median absolute deviation (MAD) of the dynamic score:

\begin{equation}
    \tau
    =
    \operatorname{median}(S)
    +
    k\,\operatorname{MAD}(S),
\end{equation}

where $k$ is a scaling factor. The dynamic segmentation mask is then

\begin{equation}
    M(u,v)
    =
    \mathbb{1}\left[S(u,v)>\tau\right].
\end{equation}

Pixels with unreliable optical-flow correspondences, such as occluded or out-of-bounds regions, are excluded from the dynamic prediction. Finally, morphological opening and closing are applied to remove isolated detections and fill small gaps in the predicted dynamic regions.

\section{Experiments and Results}

\subsection{Dynamic Segmentation Evaluation}

We evaluate the proposed radar-geometric dynamic segmentation pipeline on 10 scenes from the nuScenes dataset, comprising 394 consecutive frame pairs. Since this experiment does not involve any learned dynamic segmentation component, it provides a purely geometric baseline for evaluating the effectiveness of radar-derived ego-motion and rigid-flow consistency.

The predicted dynamic masks are evaluated against the ground-truth dynamic segmentation masks using precision, recall, F1-score, and intersection-over-union (IoU). We report pixel-level metrics aggregated across all evaluated frames, with the overall results shown in Table~\ref{tab:radar_geometric_results}. We use the 2D instance masks provided by nuInsSeg, which extends nuScenes with image-space instance segmentation annotations. We select instances associated with the nuScenes moving attribute and merge their 2D masks to obtain a binary dynamic-object mask. Evaluation is therefore performed against image-space dynamic masks rather than projections of 3D bounding boxes.

The radar-geometric baseline achieves an overall precision of $31.24\%$, recall of $36.51\%$, F1-score of $33.67\%$, and dynamic IoU of $20.24\%$. These results demonstrate that rigid-flow inconsistency induced by radar-based ego-motion provides a useful signal for identifying dynamic objects even without learning an explicit segmentation model.

However, performance varies substantially across scenes. The best performance is obtained on scene-0061, with an IoU of $36.02\%$ and an F1-score of $52.96\%$, followed by scene-0757 with an IoU of $35.28\%$ and an F1-score of $52.16\%$. In contrast, scenes such as scene-0655, scene-0916, and scene-1077 obtain IoUs below $2.1\%$. This variation indicates that the effectiveness of the geometric formulation depends strongly on scene characteristics, including the amount and distribution of dynamic content, radar coverage, depth quality, and the accuracy of the estimated ego motion. An example segmentation mask generated based on the above methodology is shown above:

\begin{figure}[htbp]
    \centering
    
    \begin{subfigure}[b]{0.45\textwidth}
        \centering
        \includegraphics[width=\textwidth]{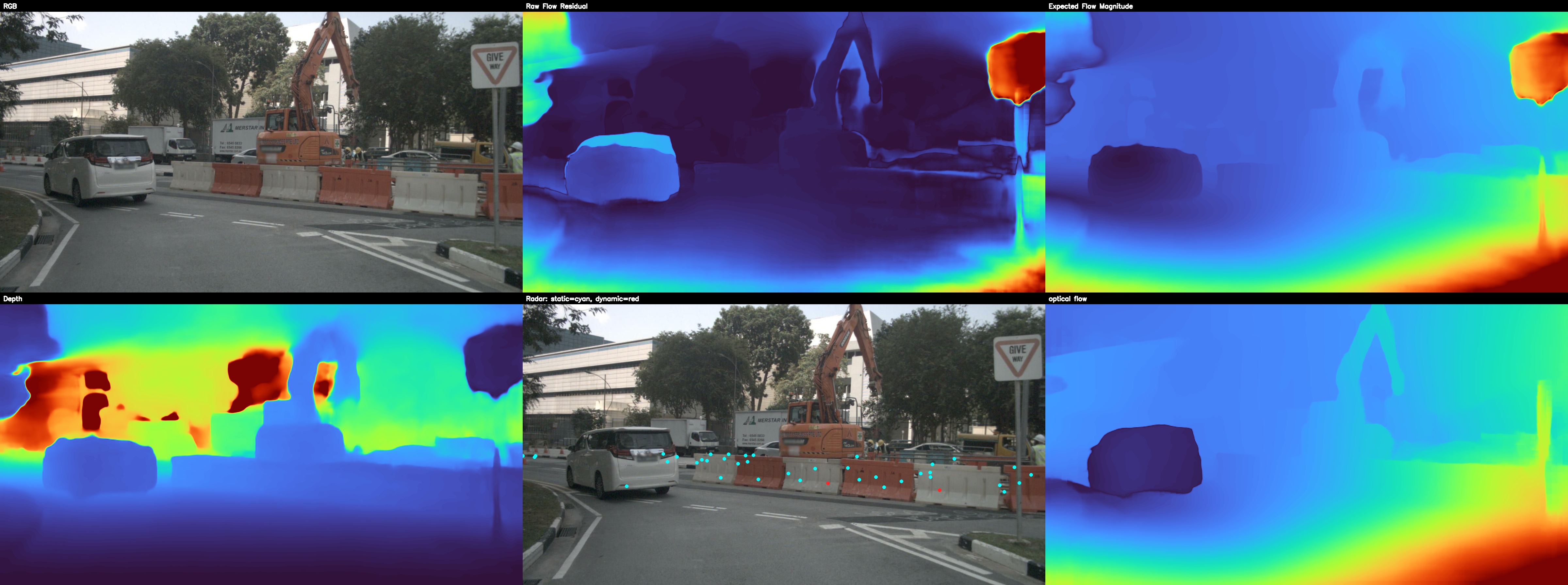}
        \caption{From Top Left to Bottom Right: RGB image, Residual between Expected and actual optical flow ($f_{obs} - f_{est}$), Expected flow based on static environment assumption ($f_{est}$), Metric depth map ($D_{t-1}$), Segmented radar pcd overlayed on image with red showing dynamic points, Flow obtained via optical flow model ($f_{obs}$)}
        \label{fig:first}
    \end{subfigure}
    \hfill 
    \begin{subfigure}[b]{0.45\textwidth}
        \centering
        \includegraphics[width=\textwidth]{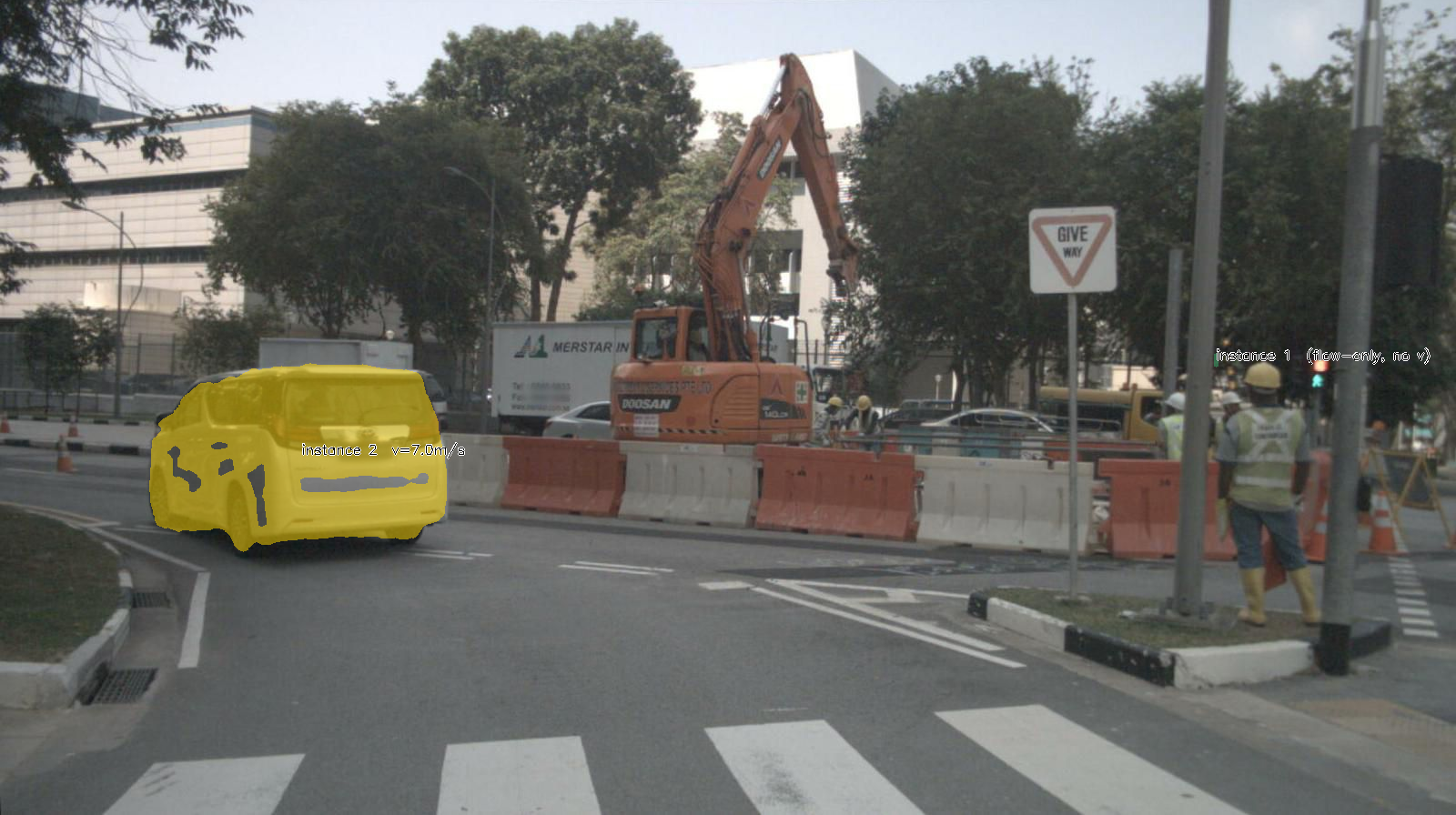}
        \caption{Final Segmentation mask.}
        \label{fig:second}
    \end{subfigure}

    \caption{Dynamic Object Segmentation based on flow consistency.} 
\end{figure}

\begin{table}[t]
\centering
\caption{Per-scene performance of the radar-geometric dynamic segmentation baseline.}
\label{tab:radar_geometric_results}
\begin{tabular}{lrrrrr}
\toprule
Scene & Frames & IoU & Precision & Recall & F1 \\
\midrule
scene-0061 & 38 & 0.3602 & 0.5177 & 0.5420 & 0.5296 \\
scene-0103 & 39 & 0.1859 & 0.2538 & 0.4097 & 0.3135 \\
scene-0553 & 40 & 0.2660 & 0.4355 & 0.4059 & 0.4202 \\
scene-0655 & 40 & 0.0201 & 0.0220 & 0.1872 & 0.0394 \\
scene-0757 & 40 & 0.3528 & 0.5243 & 0.5190 & 0.5216 \\
scene-0796 & 39 & 0.1299 & 0.1720 & 0.3464 & 0.2299 \\
scene-0916 & 40 & 0.0177 & 0.0186 & 0.2637 & 0.0348 \\
scene-1077 & 40 & 0.0165 & 0.1352 & 0.0184 & 0.0324 \\
scene-1094 & 39 & 0.2453 & 0.5561 & 0.3050 & 0.3939 \\
scene-1100 & 39 & 0.0829 & 0.2188 & 0.1177 & 0.1530 \\
\midrule
Overall & 394 & 0.2024 & 0.3124 & 0.3651 & 0.3367 \\
\bottomrule
\end{tabular}
\end{table}
\begin{figure}[h]                 
    \centering                    
    \includegraphics[width=0.5\textwidth]{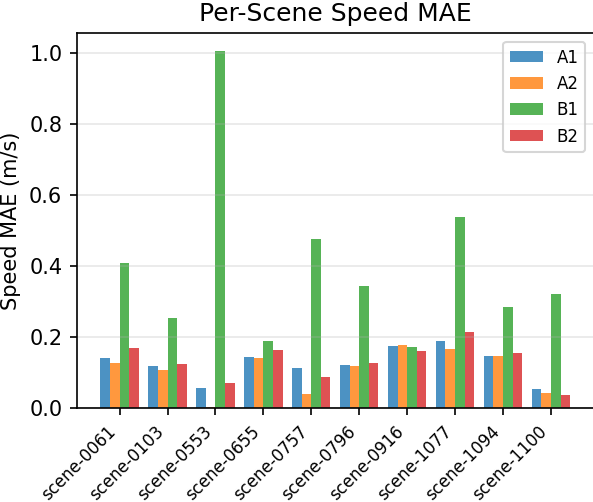} 
    \caption{Comparision of MAE across 10 scenes of Nuscenes dataset. A1 and B1 represent LSQ based solution on segmented static points and all points respectively. A2 and B2 represent LSQ+Ransac on segmented static points and all points respectively }
    \label{fig:radar_odom_MAE}          
\end{figure}

\subsection{Radar Based Velocity Estimation}

As shown in \ref{sec:radar_odometry}, ego motion can be estimated using Radar doppler returns, and further refined by segmenting the pointcloud into static and dynamic. This segmentation improves the ego velocity estimation accuracy when compared against finding velocity using the entire pointcloud.
\section{Conclusion}

In this paper, we present preliminary results for investigating the use of radar+rgb based simultaneous dynamic object and ego motion estimation. We show that the flow consistency based method can produce priors for segmenting dynamic objects in the image without any explicit learning, and can improve the ego motion estimates in a closed loop. In the future we would explore how to improve the segmentation results and incorporate tracking by incorporating semantic information from RGB image and also on how to integrate Radar based velocity estimation with a multimodal sensor based localization pipeline. 

\bibliographystyle{IEEEtran}
\bibliography{references}

\end{document}